\documentclass[10pt,twocolumn,letterpaper]{article}

\usepackage{cvpr}
\usepackage{times}
\usepackage{epsfig}
\usepackage{graphicx}
\usepackage{amsmath}
\usepackage{amssymb}

\graphicspath{{figure/}}

\usepackage{setspace}
\usepackage{mathtools}

\usepackage{algorithmicx}
\usepackage{algpseudocode}
\usepackage{algorithm}

\usepackage{enumitem}
\usepackage{xspace}
\usepackage{xparse}

\usepackage{csvsimple}

\usepackage[pagebackref=true,breaklinks=true,letterpaper=true,colorlinks,bookmarks=false]{hyperref}

\cvprfinalcopy 

\def\cvprPaperID{3577} 

\newtheorem{lemma}{Lemma}
\newtheorem{theorem}{Theorem}

\newcommand{\ud}{\,\mathrm{d}}

\newcommand{\R}{\mathbb{R}}

\newcommand{\comment}[1]{ }

\newcommand{\argmin}{\arg\!\min}
\newcommand{\prox}{\textrm{prox}}

\DeclareDocumentCommand\newstep{o}{%
\item\IfNoValueTF{#1}{}{#1 \textendash\xspace}}

\newlist{steps}{enumerate}{1}
\setlist[steps]{label=\textit{Step \arabic*:},leftmargin=*}

\begin{document}

\title{Adaptive Regularization in Convex Composite Optimization\\ for Variational Imaging Problems}

\author{Byung-Woo Hong \qquad Ja-Keoung Koo\\
Chung-Ang University, Korea\\
{\tt\small \{hong,jakeoung\}@cau.ac.kr}
\and
Hendrik Dirks \qquad Martin Burger\\
University of M{\"u}nster, Germany\\
{\tt\small \{hendrik.dirks,martin.burger\}@wwwu.de}
}

\maketitle

\begin{abstract}

We propose an adaptive regularization scheme in a variational framework where a convex composite energy functional is optimized. We consider a number of imaging problems including denoising, segmentation and motion estimation, which are considered as optimal solutions of the energy functionals that mainly consist of data fidelity, regularization and a control parameter for their trade-off. We presents an algorithm to determine the relative weight between data fidelity and regularization based on the residual that measures how well the observation fits the model. Our adaptive regularization scheme is designed to locally control the regularization at each pixel based on the assumption that the diversity of the residual of a given imaging model spatially varies. The energy optimization is presented in the alternating direction method of multipliers (ADMM) framework where the adaptive regularization is iteratively applied along with mathematical analysis of the proposed algorithm. We demonstrate the robustness and effectiveness of our adaptive regularization through experimental results presenting that the qualitative and quantitative evaluation results of each imaging task are superior to the results with a constant regularization scheme. The desired properties, robustness and effectiveness, of the regularization parameter selection in a variational framework for imaging problems are achieved by merely replacing the static regularization parameter with our adaptive one.

\end{abstract}

\section{Introduction}

A variety of computer vision problems can be casted as energy minimization problems in a variational framework where an energy functional is formulated and the minimum energy is attained at the solution to the problem. One fundamental categorization of the energy functional is convex or non-convex. The advantage of convex energy is that a unique global solution can be obtained independent of the initial condition in contrast to non-convex energy that may have several local minima. Although the non-convex formulation often accounts for more realistic imaging models~\cite{nikolova2010fast,hintermuller2013nonconvex,ochs2013iterated,mollenhoff2015primal}, the desirable computational property of convex formulations has led to recent advances in their efficient optimization algorithms~\cite{nesterov1994interior,beck2009fast,goldstein2009split,esser2010general,chambolle2011first,becker2011nesta,combettes2011proximal,parikh2014proximal,goldstein2014fast}. Such convex optimization techniques have been applied to various computer vision problems including image denoising~\cite{chan2006optimization,zibulevsky2010l1,setzer2011operator}, segmentation~\cite{chan2006algorithms,pock2011diagonal,chambolle2012convex} and motion estimation~\cite{wedel2009improved,ayvaci2010occlusion,unger2012joint}.
The convex optimization of such problems in a variational approach generally has the composite form of a data fidelity term and a regularization term. The data fidelity term measures the discrepancy between observation and model, whereas the regularization term incorporates additional a-priori information about the solution. The trade-off between the model fit and the regularity is usually controlled by a static positive weight. This parameter is often critically related to the quality of the solution. One of the common criteria for determining suitable values of the control parameters in a variational framework is the manual selection via extensive visual inspections or the exhaustive search with respect to certain quality measures via a training process. In addition to the difficulty and sensitivity of selecting an optimal control parameter, the static trade-off between the data fidelity and the regularization is not suited for considering intermediate solutions that are led to better final solution with an alternative adaptation of the balancing between the two terms in the optimization procedure where the data fidelity and the regularization energies keep changing for a balance. Another aspect of the need for adaptive regularization parameter is that it is desirable to consider local residual that is related to the degree of desired regularity of the solution. For example, a constant global regularization parameter is not effective to cope with multiple objects with different velocities in the motion estimation application. Similarly, a constant global regularity often fails to deal with spatially varying noises in the image denoising or segmentation problem. 
In this work, we propose a novel algorithm for adjusting the regularization parameter that is locally determined by the intermediate solution at each iteration of the optimization procedure. The iterative adaptation of the regularization parameter between the data fidelity and the regularization facilitates the optimization process to obtain more precise results. In addition to the dynamic property of the proposed adaptive regularization, we consider local residual for adaptively determining the degree of regularization in order to deal with statistical discrepancy that may spatially vary between the model and observation.

\subsection{Related Work}

For image denoising problems, the noise variation has been estimated for choosing the proper value of the regularization parameter in~\cite{galatsanos1992methods} and the stability analysis of the parameter estimate has been performed in~\cite{thompson1991study}. 
As a selection criterion of the regularization parameter, the generalized cross-validation has been used for image restoration applications in~\cite{nguyen2001efficient}. 
Another alternative has been proposed to use plots of the norm of the solution versus the norm of the regularity, called $L$-curve, for the criterion of the regularization parameter selection in~\cite{mc2003direct}. The truncated singular value decomposition~\cite{watzenig2004adaptive}, $U$-curve~\cite{krawczyk2007regularization}, and generalization of the maximum likelihood estimate~\cite{wahba1985comparison} have also been proposed in determining the global regularity.
To infer the value of regularization parameters from the observed data, a number of techniques have been proposed. In the computation of the optical flow, the regularization parameter is chosen in such a way that the estimated error is minimized~\cite{ng1997data}, and the joint probability of the gradient field and the velocity field is maximized~\cite{krajsek2006maximum}. Another approach is to apply a smoothing kernel on the approximated flow using bilateral filtering~\cite{lee2010optical} and incorporate noise estimation~\cite{chantas2014variational}. A non-local regularization has been applied for the computation of motion in~\cite{werlberger2010motion,krahenbuhl2012efficient,ranftl2014non}. The image gradient information has been widely used in the form of edge indicator function as a weighting factor to the regularity in the computation of optical flow~\cite{werlberger2009anisotropic,wedel2009structure}. Moreover, this technique has also been applied to the image segmentation problem~\cite{caselles1997geodesic,bresson2007fast}. Alternatively, a learning scheme has been used to measure the segmentation quality with AdaBoost where the optimal regularization parameter is selected with respect to the learned measures in~\cite{peng2008parameter}.
In most computer vision problems, static information from the observation is considered to control regularization. The static regularity is often ineffective to guide the optimization procedure due to the spatially varying residual diversity, which is a motivation to propose an iterative regularity scheme with spatial adaptation.
%

\section{Variational Model with Adaptive Regularization}

Let $\mathcal{U}$ and $\mathcal{W}$ be finite dimensional real vector spaces equipped with inner product $\langle \cdot, \cdot \rangle$ and induced norm $\| \cdot \|$. Let $K : \mathcal{U} \rightarrow \mathcal{W}$ be a continuous linear operator with the induced norm:
\begin{align}
\| K \| = \max \{ \| Ku \| : u \in \mathcal{U}, \, \| u \| \le 1 \}.
\end{align}
Our problem of interest is to solve the following composite convex optimization formulation:
\begin{align}
\min_{u \in \mathcal{U}} \lambda \, \mathcal{D}(u) + (1 - \lambda) \, \mathcal{R}(Ku),
\end{align}
where $\lambda \in [0, 1]$ is a control parameter that determines the overall trade-off between the two terms $\mathcal{D} : \mathcal{U} \rightarrow \R$ and $\mathcal{R} : \mathcal{W} \rightarrow \R$. The functionals $\mathcal{D}$ and $\mathcal{R}$ are assumed to be closed, proper and convex. 
The control parameter $\lambda$ determines the relative weight of the two terms in the objective functional in which the functional $\mathcal{D}$ generally corresponds to the data fidelity and the functional $\mathcal{R}$ to the regularization. 
In most cases, $\lambda$ takes a constant value over the entire domain of the unknown function $u$, thus it considers the overall weight between the data fidelity and the regularization. However, the constant control parameter does not take into account the local balance of the two terms, which may occur with spatially varying residual of the image model. 
Thus, we propose to apply a spatially adaptive regularization that locally determines the relative weight based on the local fit of observation to the model. 
We now consider the objective functional $\mathcal{E}_\lambda$ that consists of the data fidelity $\rho$ and the regularization $\gamma$ with a spatially adaptive weighting parameter $\lambda$:
%
\begin{align} \label{eq:adaptive}
\mathcal{E}_\lambda(u) &= \int_{\Omega} \lambda \, \rho(u) \ud x + \int_{\Omega} (1 - \lambda) \, \gamma(Ku) \ud x,\\
\lambda &= \exp \left( - \frac{\rho(u)}{\beta} \right),
\end{align}
where $\Omega$ denotes the domain of the unknown function $u$, and the parameter $\beta \in \R$ is related to the distribution of the values in $\lambda$ that are restricted to the range $(0, 1]$.
The adaptive weighting parameter $\lambda$ is designed to use higher weights for the points in the data fidelity where the residual $\rho$ is lower so that the regularization $\gamma$ is less imposed. On the other hand, lower weights are applied to the points in the data fidelity where the residual $\rho$ is higher so that the regularization $\gamma$ is more imposed. As we shall see below, the model in~\eqref{eq:adaptive} has a certain bias towards achieving $\rho(u)=0$, hence in some applications it will be beneficial to use the following model:
\begin{align} \label{eq:adaptiveeps}
\mathcal{E}_\lambda(u) &= \int_{\Omega} (\lambda \, \rho(u) +(1 - \lambda) \, \gamma(Ku) ) \ud x,\\
\lambda &= (1- \epsilon) \exp \left( - \frac{G*\rho(u)}{\beta} \right),
\end{align}
with some small $\epsilon > 0$ to ensure that $\lambda(u)$ is actually positive, hence there is nonzero regularization over the entire domain and therefore well-posedness holds. 
The additional convolution with a kernel $G$, e.g. a Gaussian with small variance, can be applied to promote smoothness in the regularization parameter. Note that the original model in~\eqref{eq:adaptive} can be understood as the special case of $\epsilon = 0$ and $G$ being the Dirac delta in the model in~\eqref{eq:adaptiveeps}.
Note that we only look for a minimizer of $\mathcal{E}_\lambda$ for fixed $\lambda$ meaning that we are not jointly minimizing $\mathcal{E}_{\lambda(u)}(u)$ with respect to $u$. This minimization procedure is considered as a natural fixed-point map $u \mapsto \lambda \mapsto \argmin \mathcal{E}_\lambda$ of the mathematical structure in~\eqref{eq:adaptive}, respectively in~\eqref{eq:adaptiveeps}. Its numerical solution as well as the mathematical analysis will be provided using a fixed point problem framework in the following sections.

%
%
\subsection{Optimization with ADMM Algorithm}

In the computation of optimal $u$ in~\eqref{eq:adaptive}, we provide an optimization scheme in the framework of the alternating direction method of multipliers (ADMM) algorithm~\cite{Boyd2010}. The optimization problem of the objective functional $\mathcal{E}_\lambda$ in~\eqref{eq:adaptive} is represented by the splitting of variables with a new variable $z = K u$ as follows:
%
\begin{align} \label{eq:objective_splitting}
\min_{u, z} \langle \lambda, \rho(u) \rangle &+ \langle 1 - \lambda, \gamma(z) \rangle \; \text{ subject to } \; z = K u,\\
\lambda &= \exp \left( - \frac{\rho(u)}{\beta} \right),
\end{align}
where $K$ is a continuous linear operator, $\langle \cdot, \cdot \rangle$ denotes the inner product, and $\beta > 0$ is a scalar parameter. 
The associated augmented Lagrangian with~\eqref{eq:objective_splitting} in the scaled form~\cite{Boyd2010} is given as:
\begin{align} \label{eq:lagrange_objective_splitting}
\mathcal{L}_{\mu}(u, z, y) = \langle \lambda, \rho(u) \rangle + \langle 1 - \lambda, \gamma(z) \rangle + \frac{\mu}{2} \| K u - z + y \|_2^2,
\end{align}
where $\mu > 0$ is a scalar augmentation parameter, and $y$ is a Lagrangian multiplier associated with $u$ and $z$.
The ADMM algorithm is an alternative minimization scheme that consists in minimizing the augmented Lagrangian in~\eqref{eq:lagrange_objective_splitting} with respect to the primal variables $u$ and $z$, and applying a gradient ascend scheme to the dual variable $y$. The update of the adaptive regularization parameter $\lambda$ is followed by the update of the variables $u$, $z$ and $y$. The optimization procedure using ADMM algorithm is presented in Algorithm \ref{alg:admm_general} where $k$ is the iteration counter.
\begin{algorithm}[tb]
\caption{The ADMM updates for optimizing~\eqref{eq:objective_splitting}}
\label{alg:admm_general}
\begin{algorithmic}
\State
\begin{align}
\hspace{-5pt} u^{k+1} & :=  \argmin_u \langle \lambda^k, \rho(u) \rangle + \frac{\mu}{2} \| K u - z^k + y^k \|_2^2 \label{step:u_general}\\
\hspace{-5pt} z^{k+1} &:= \argmin_z \langle 1 - \lambda^k, \gamma(z) \rangle + \frac{\mu}{2} \| K u^{k+1} - z + y^k \|_2^2 \label{step:z_general}\\
\hspace{-5pt} y^{k+1} &:= y^k + K u^{k+1} - z^{k+1} \label{step:y_general}\\
\hspace{-5pt} \lambda^{k+1} &:= \exp \left( - \frac{\rho(u^{k+1})}{\beta} \right) \label{step:l_general}
\end{align}
\end{algorithmic}
\end{algorithm}
The optimality condition of the update for the primal variable $u^{k+1}$ in~\eqref{step:u_general} can be simplified by the linearization of the quadratic regularization term using the Taylor expansion at around $u^k$ in combination with an additional quadratic regularity as follows:
%
\begin{align}
u^{k+1} := & \argmin_u \langle \lambda^k, \rho(u) \rangle + \mu K^* ( K u^k - z^k + y^k ) u \nonumber\\
&+ \frac{\tau}{2} \| u - u^k \|_2^2, \label{step:u_linear_general}
\end{align}
where $K^*$ denotes the adjoint operator of $K$, and $\tau > 0$ is a scalar regularity parameter.
Then, the optimality condition for the update of the primal variables $u^{k+1}$ and $z^{k+1}$ yields:
\begin{align}
\hspace{-5pt}0 &\in \lambda^k \, \partial \rho(u) + \mu K^* ( K u^k - z^k + y^k ) + \tau (u - u^k),
\label{eq:optimality_u}\\
\hspace{-5pt}0 &\in (1 - \lambda^k) \, \partial \gamma(z) - \mu (K u^{k+1} - z + y^k), \label{eq:optimality_z}
\end{align}
where $\partial$ denotes the subdifferential operator.
The solutions for updating $u$ in~\eqref{eq:optimality_u} and $z$ in~\eqref{eq:optimality_z} are obtained by the proximal operator:
\begin{align}
\hspace{-5pt}u^{k+1} &:= \prox \left( u^k - \frac{\mu}{\tau} K^* (K u^k - z^k + y^k ) \left| \frac{\lambda^k}{\tau} \right. \rho \right) \label{eq:prox_u}\\
\hspace{-5pt}z^{k+1} &:= \prox \left( K u^{k+1} + y^{k} \left| \frac{1 - \lambda^k}{\mu} \right. \gamma \right), \label{eq:prox_z}
\end{align}
where the proximal operator is defined by:
\begin{align}
\prox(v \, | \, \mu f) := \argmin_x \left( \frac{1}{2} \| x - v \|_2^2 + \mu f(x) \right), 
\label{eq:prox}
\end{align}
where $\mu > 0$ is the weighting parameter. The solution for the proximal operator $\prox(v \, | \, \mu f)$ of the $L_1$ norm when $f(x) = \| x \|_1$ is obtained by the soft shrinkage operator $S(x \, | \, \mu)$ with threshold $\mu$ as defined by:
\begin{align}
S(x \, | \, \mu) &= 
\begin{cases}
x - \mu & \quad : x > \mu\\
0 & \quad : \| x \|_1 \le \mu\\
x + \mu & \quad : x < -\mu\\
\end{cases}
\label{eq:shrink}
\end{align}

%
\subsection{Mathematical Analysis}

In the following section, we further comment on the mathematical structure of the model in~\eqref{eq:adaptive}, respectively in~\eqref{eq:adaptiveeps}.
Let us first give some motivation of the data term from the information theoretic perspective. Assume we want to choose some optimal parameter $\lambda$ in a model of the form
$$ u \in \argmin \left( \int_\Omega \lambda(x) \rho(u(x)) \ud x + \mathcal{R}(u)\right), $$
where $\mathcal{R}$ is a given regularization functional. Then we can look for a parameter that maximizes an approximate entropy $-\lambda \log \lambda+\lambda-1$, which is related to maximizing its information content. When penalizing the given approximate entropy measure (note that we minimize the negative entropy) we arrive at 
\begin{align}
(u,\lambda) \in \argmin & \left( \int_\Omega \lambda(x) \rho(u(x)) \ud x + \mathcal{R}(u) \right. \nonumber \\
& + \left. \beta \int_\Omega (\lambda(x) \log \lambda(x) - \lambda(x) + 1 ) \ud x\right), \nonumber
\end{align}
where $\beta$ is a positive scalar parameter.
Another interpretation of the additional entropy terms is a Kullback-Leibler divergence between $\lambda$ and the constant parameter equal to one, i.e. with large $\beta$ we penalize arbitrary local changes with respect to a constant parameter. It is straightforward to see that a minimizer $\lambda$ from the above minimization is the same as the one in~\eqref{eq:adaptive}. 
%
%
%
%
%
A first general property that explains the behavior of the model in~\eqref{eq:adaptiveeps} at $\epsilon = 0$ is the following:
\begin{lemma}
	Assume there exists $u^*$ with $\rho(u^*) \equiv 0$, then $u^*$ is a fixed point of~\eqref{eq:adaptive}, respectively of~\eqref{eq:adaptiveeps} for $\epsilon = 0$.
\end{lemma}
The proof is provided in the supplementary material.
In some cases such as a denoising problem, this property may be undesirable, since choosing $u^*$ equal to the noisy image always yields a solution, hence $\epsilon > 0$ introduces a necessary minimal regularization. In other examples such as the segmentation problem based on a piecewise constant image model this property turns out to be useful, since it can lead to a precise segmentation solution from a piecewise constant observation. On the other hand, an image with local variations $\rho(u)$ being non-zero everywhere leads to an immediate regularization.
We now provide a brief well-posedness analysis for the proposed model in the following. For this sake, we consider a particular case of $K=\nabla$ and the space to be minimized on being $BV(\Omega)$ for $\Omega \subset \R^d$ a bounded domain. The correct mathematical definition of the regularization functional is then given by
$$ \mathcal{R}_\lambda(u) = \sup_{\varphi \in C_0^\infty(\Omega;\R^d), \Vert \varphi \Vert_\infty \leq 1} 
\int_\Omega u \nabla \cdot ( (1-\lambda) \varphi) \ud x. $$
The problem we consider is then the minimization of the following:
%
\begin{align} \label{eq:adaptiveentropy}
	E_\lambda(u) &= \int_{\Omega} \lambda  \, \rho(u) \ud x + \mathcal{R}_{\lambda}(u) ,\\
	\lambda &= (1- \epsilon) \exp \left( - \frac{G*\rho(u)}{\beta} \right).
\end{align}
For ease of mathematical presentation, we consider the minimization on the space of functions $u$ of bounded variation with mean zero, which we  denote by $BV_0(\Omega)$.
In order to verify the existence of a solution for our model, it is natural to consider the fixed point map 
$ u \mapsto \lambda \mapsto u = \argmin E_\lambda$, from which we derive the following theoretical result:
\begin{theorem}
	Let $\epsilon > 0$ be arbitrary and $\beta > 0$ be sufficiently large. Let $G$ be bounded, integrable, and continuously differentiable with bounded and integrable gradient. Moreover, let $\rho$ be a continuous, nonnegative, convex functional, such that the minimizer of $E_\lambda$ is unique for every $\lambda$.  Then there exists a fixed-point $u \in BV_0(\Omega)$ for~\eqref{eq:adaptiveentropy}
\end{theorem} 
The detailed mathematical proof is provided in the supplementary material. It is also noted that the uniqueness of $E_\lambda$ in~\eqref{eq:adaptiveentropy} for fixed $\lambda$ is guaranteed if $\rho$ is strictly convex. 

%
\section{Applications} \label{sec:application}

We demonstrate the effectiveness and robustness of the proposed variational model that incorporates the adaptive regularization scheme in the application of image denoising, image segmentation and motion estimation. We present the variational energy formulations for those problems using the classical models where the constant regularization parameter is replaced with our adaptive one. The details of the optimization algorithm based on the ADMM method for each imaging task are also provided in the supplementary material. 

%
\subsection{Image Denoising}

We consider an image denoising problem as a constrained total variation minimization problem using the classical model proposed in~\cite{rudin1992nonlinear}. Let $f : \Omega \rightarrow \R$ be an input noisy image and $u : \Omega \rightarrow \R$ be its reconstruction based on the following energy functional:
%
\begin{align} \label{eq:energy_denoise}
\min_{u} & \int_{\Omega} \lambda \frac{(u - f)^2}{2} \ud x + \int_{\Omega} (1 - \lambda) \| \nabla u \|_1 \ud x,\\
\lambda &= \exp \left( - \frac{(u - f)^2}{2 \beta} \right),
\end{align}
where $\lambda$ denotes the proposed adaptive regularization parameter that is determined based on the data fidelity term, and $\beta > 0$ is a scalar parameter related to the overall diversity of the residual. The adaptive regularization parameter $\lambda$ considers the local residual that measures the discrepancy between measurement $f$ and reconstruction $u$. This adaptivity leads to a desirable solution in particular when spatially biased degradation factors such as noises exist.

%

%

\subsection{Image Segmentation}

We consider an image segmentation problem based on the piecewise constant model~\cite{chan2001active}. Let $f : \Omega \rightarrow \R$ be an input image. A convex energy formulation for a bi-partitioning problem~\cite{bresson2007fast} with the proposed adaptive regularization parameter $\lambda$ leads to:
\begin{align} \label{eq:energy_const}
\min_{0 \le u \le 1} & \int_{\Omega} \lambda \left\{ (f - c_1)^2 u + (f - c_2)^2 (1-u) \right\} \ud x \nonumber\\
&+ \int_{\Omega} (1 - \lambda) \| \nabla u \|_1 \ud x,
\end{align}
where $c_1$ and $c_2$ are estimates of the interior and exterior of the segmenting boundary, respectively. A smooth function $u : \Omega \rightarrow [0, 1]$ represents a partitioning interface that determines regions $\Omega_1$ and $\Omega_2$ by thresholding with a parameter $\theta \in [0, 1]$:
\begin{align}
\Omega_1 = \{ x \in \Omega | u(x) > \theta \}, \quad \Omega_2 = \{ x \in \Omega | u(x) \le \theta \}, \nonumber
\end{align}
where $\Omega_1 \cup \Omega_2 = \Omega$, and the usual choice of the threshold is $\theta = 0.5$.
The adaptive regularization parameter $\lambda$ is determined by the data fidelity term:
\begin{align} \label{eq:lambda_segmentation}
\lambda = \exp \left( - \frac{(f - c_1)^2 \, u + (f - c_2)^2 \, (1 - u)}{\beta} \right),
\end{align}
where $\beta > 0$ is a scalar parameter.
The constraint on $u(x) \in [0, 1]$ can be imposed by adding an indicator function $\delta_{\mathcal{C}}$ on the convex set $C = [0, 1]$ to the objective functional:
\begin{align} \label{eq:indicator}
\delta_{\mathcal{C}} (u) =
	\begin{cases}
	0 & \quad \text{if } u \in \mathcal{C}\\
	\infty & \quad \text{if } u \not\in \mathcal{C}.
	\end{cases}
\end{align}
Then, the unconstrained objective functional reads:
\begin{align*} 
\min_u & \int_{\Omega} \lambda \left\{ (f - c_1)^2 u + (f - c_2)^2 (1-u) \right\} \ud x
\\
&+ \int_{\Omega} (1 - \lambda) \| \nabla u \|_1 \ud x + \delta_{\mathcal{C}} (u),
\end{align*}
and it reduces to:
\begin{align} \label{eq:energy_simple}
\min_u \int_{\Omega} \lambda(u) \, \rho(u) \ud x + \int_{\Omega} (1 - \lambda(u)) \, \gamma(\nabla u) \ud x + \delta_{\mathcal{C}} (u),
\end{align}
where $\rho(u) = \left( (f - c_1)^2 - (f - c_2)^2 \right) u$ and $\gamma(\nabla u) = \| \nabla u \|_1$. The regularization parameter $\lambda$ is designed to adaptively choose regularity depending on the residual $\rho$. A higher residual would allow more regularity to effectively deal with inhomogeneity. In the same way, a lower residual would impose a higher regularity so that a precise segmentation boundary can be obtained despite its complex shape that is often unnecessarily blurred due to an undesirable uniform regularity over the entire image domain. 

%

%
\subsection{Motion Estimation}

For the motion estimation problem, we let $I_0(x), I_1(x) : \Omega \rightarrow \R$ be images taken at two different time instances and $\Omega$ be the image domain. We consider an optical flow model based on the brightness consistency assumption \cite{horn1981determining}:
\begin{align}
I_1(x+\boldsymbol{v}(x)) - I_0(x) = 0.
\end{align}
Due to the non-linearity of the formulation above, one can linearize the first term with respect to some a-priori solution $\boldsymbol{v}_0$ close to $\boldsymbol{v}$ leading to:
\begin{align}
\rho(\boldsymbol{v}(x)) := &\nabla I_1(x+\boldsymbol{v}_0(x))\cdot (\boldsymbol{v}(x) - \boldsymbol{v}_0(x)) \nonumber\\
&+ I_1(x+\boldsymbol{v}_0(x)) - I_0(x).
\end{align}
The linearized brightness consistency assumption can then be used as a data fidelity term together with a total variation regularization on both components of the velocity field (see \cite{wedel2009improved,perez2013tv}) leading to the following convex energy formulation:
\begin{align} \label{eq:energy_motion}
\int_{\Omega} \lambda(x) \,  | \rho(\boldsymbol{v}(x)) |  + \sum_{i=1}^{2} (1 - \lambda(x)) \, \| \nabla v_i(x) \|_2 \ud x,
\end{align}
where $i$ indicates the component index, and $\lambda(x)$ denotes the proposed adaptive regularization parameter:
\begin{align} \label{eq:lambda_motion}
\lambda(x) = \exp \left( - \frac{| \rho(\boldsymbol{v}(x)) |}{\beta} \right),
\end{align}
where $\beta > 0$ is a scalar parameter related to the overall distribution of $0 < \lambda \le 1$. Since the optical flow constraint is valid for small motion, we consider a coarse-to-fine approach including warping on each level of the pyramid. As shown in~\eqref{eq:lambda_motion}, the regularization parameter $\lambda$ automatically controls the trade-off between the data fidelity and the regularization, imposing a higher regularity on the velocity where mismatch occurs, for example at occlusions. On the other hand, it is unnecessary to impose any regularity on the velocity where perfect match is achieved. 

%

%
\section{Numerical Results}
%
%
%
%
%
%
%
\def\fH{60pt}
\def\case{124084}
\begin{figure*}[tb]
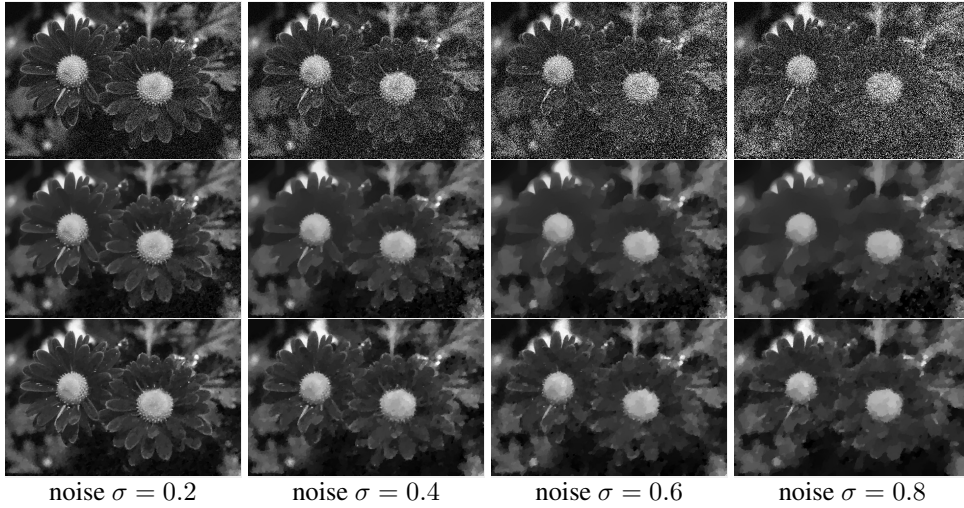

\centering
\begin{tabular}{c@{ }c@{ }c@{ }c}
\includegraphics[height=\fH]{denoise/berkeley/\case/{{img_noisy_0.20}}} &%
\includegraphics[height=\fH]{denoise/berkeley/\case/{{img_noisy_0.40}}} &%
\includegraphics[height=\fH]{denoise/berkeley/\case/{{img_noisy_0.60}}} &%
\includegraphics[height=\fH]{denoise/berkeley/\case/{{img_noisy_0.80}}}\\[-\dp\strutbox] 
\includegraphics[height=\fH]{denoise/berkeley/\case/{{img_denoised_best_psnr_constant_std_noise_0.20}}} &%
\includegraphics[height=\fH]{denoise/berkeley/\case/{{img_denoised_best_psnr_constant_std_noise_0.40}}} &%
\includegraphics[height=\fH]{denoise/berkeley/\case/{{img_denoised_best_psnr_constant_std_noise_0.60}}} &%
\includegraphics[height=\fH]{denoise/berkeley/\case/{{img_denoised_best_psnr_constant_std_noise_0.80}}}\\[-\dp\strutbox] 
\includegraphics[height=\fH]{denoise/berkeley/\case/{{img_denoised_best_psnr_adaptive_std_noise_0.20}}} &%
\includegraphics[height=\fH]{denoise/berkeley/\case/{{img_denoised_best_psnr_adaptive_std_noise_0.40}}} &%
\includegraphics[height=\fH]{denoise/berkeley/\case/{{img_denoised_best_psnr_adaptive_std_noise_0.60}}} &%
\includegraphics[height=\fH]{denoise/berkeley/\case/{{img_denoised_best_psnr_adaptive_std_noise_0.80}}}\\[-\dp\strutbox] 
noise $\sigma = 0.2$ & noise $\sigma = 0.4$ & noise $\sigma = 0.6$ & noise $\sigma = 0.8$
\end{tabular}
\caption{[Denoising] Visual comparison of the denoising results with the best PSNR. 
(top) input images with spatially biased Gaussian noises with different standard deviations. 
(middle) optimal solutions using the constant regularity.
(bottom) optimal solutions using our adaptive regularity.}
\label{fig:denoise_image}
\end{figure*}
%
%
%
\def\fH{120pt}
\def\case{124084}
\begin{figure*}
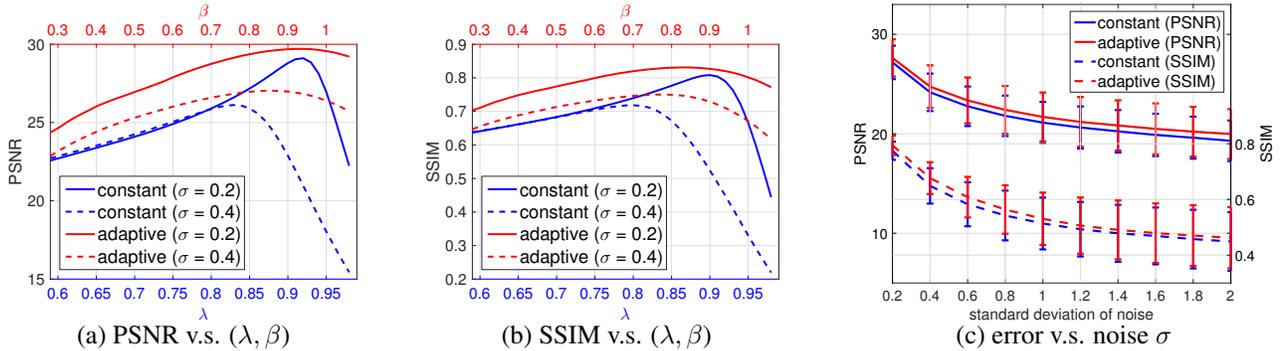

\centering
\begin{tabular}{c@{ }c@{ }c}
\includegraphics[height=\fH]{denoise/berkeley/\case/{{psnr_vs_params}}} \hspace{10pt} & \hspace{10pt}%
\includegraphics[height=\fH]{denoise/berkeley/\case/{{ssim_vs_params}}} \hspace{10pt} & \hspace{10pt} %
\includegraphics[height=\fH]{denoise/berkeley/{{errorbar_using_std}}}\\[-\dp\strutbox] 
(a) PSNR v.s. ($\lambda, \beta$) \hspace{10pt} & \hspace{10pt} (b) SSIM v.s. ($\lambda, \beta$) \hspace{10pt} & \hspace{10pt} (c) error v.s. noise $\sigma$
\end{tabular}
\caption{[Denoising] Error measures with varying regularization parameters for (a) PSNR and (b) SSIM, where the bottom x-axis represents $\lambda$ (constant) and the top x-axis represents $\beta$ (adaptive). (c) PSNR (left y-axis) and SSIM (right y-axis) for the images with varying noise standard deviations (x-axis).}
\label{fig:denoise_error_param}
\end{figure*}
%

In the following experiments, we demonstrate the robustness and effectiveness of our proposed adaptive regularization scheme in the application of image denoising, image segmentation and motion estimation. 
The major objective of the following experiments is to present the advantage of using the adaptive regularization algorithm over the conventional static one. Thus, we use the classical model for each problem as shown in the previous section and compare the performance of the same algorithm using our adaptive scheme against the static one. Note that the adaptive regularization scheme can be integrated into more sophisticated models by merely replacing their regularization parameter with our adaptive one based on the residual of the model under consideration.
%
In the experiments, we denote by $\lambda \in \R$ the control parameter for the conventional static regularization and by $\beta \in \R$ for our adaptive regularization in~\eqref{eq:adaptive}. The same values for the common parameters in the ADMM optimization are used $\mu = 1$ and $\tau = 8$ (see the supplementary material for more details of the optimization algorithm).

%
%
%
\subsection{Image Denoising}
We solve the energy minimization problem in~\eqref{eq:energy_denoise} for the restoration of noisy images with spatially biased Gaussian noises of different noise levels. For the quantitative evaluation, we measure the structural similarity (SSIM) index~\cite{wang2004image} and the peak signal-to-noise ratio (PSNR). In Fig.~\ref{fig:denoise_image}, the best results with respect to PSNR\footnote{The best denoising results with respect to SSIM are similar to the ones with PSNR, thus we only present the results with PSNR.} for the input images (top) with different noise standard deviations $\sigma$ using the constant regularization parameter (middle) and our adaptive one (bottom) are presented. The results with the constant regularization parameter indicate that undesired excessive smoothing is globally applied to cope with the highest noise level that is locally present. Whereas, the results with our adaptive regularity parameter demonstrate that the degree of smoothing is locally determined by the spatially varying noise levels.
%
%
%
%
%
\def\fH{45pt}
\def\case{353013}
\begin{figure*}[htb]
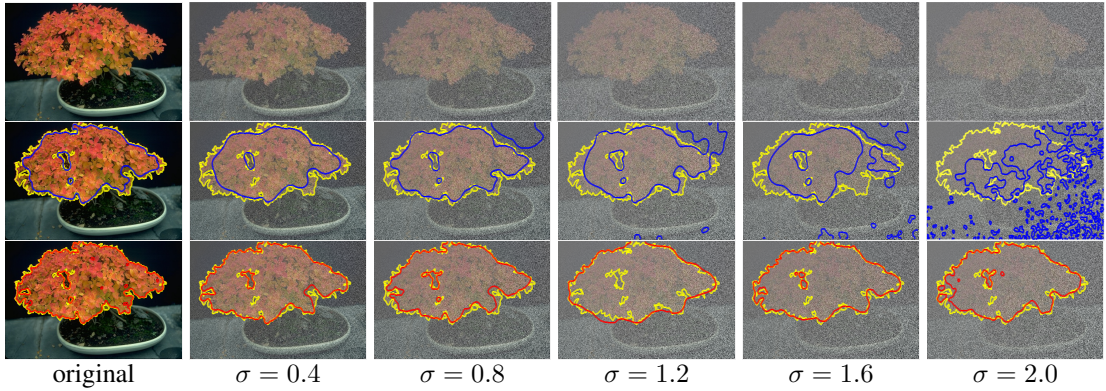

\centering
\begin{tabular}{c@{ }c@{ }c@{ }c@{ }c@{ }c}
\includegraphics[height=\fH]{segmentation/berkeley/\case/{{img_noisy_0.00}}}&%
\includegraphics[height=\fH]{segmentation/berkeley/\case/{{img_noisy_0.40}}}&%
\includegraphics[height=\fH]{segmentation/berkeley/\case/{{img_noisy_0.80}}}&%
\includegraphics[height=\fH]{segmentation/berkeley/\case/{{img_noisy_1.20}}}&%
\includegraphics[height=\fH]{segmentation/berkeley/\case/{{img_noisy_1.60}}}&%
\includegraphics[height=\fH]{segmentation/berkeley/\case/{{img_noisy_2.00}}}\\[-\dp\strutbox] 
\includegraphics[height=\fH]{segmentation/berkeley/\case/{{contour_constant_std_noise_0.00}}}&%
\includegraphics[height=\fH]{segmentation/berkeley/\case/{{contour_constant_std_noise_0.40}}}&%
\includegraphics[height=\fH]{segmentation/berkeley/\case/{{contour_constant_std_noise_0.80}}}&%
\includegraphics[height=\fH]{segmentation/berkeley/\case/{{contour_constant_std_noise_1.20}}}&%
\includegraphics[height=\fH]{segmentation/berkeley/\case/{{contour_constant_std_noise_1.60}}}&%
\includegraphics[height=\fH]{segmentation/berkeley/\case/{{contour_constant_std_noise_2.00}}}\\[-\dp\strutbox]
\includegraphics[height=\fH]{segmentation/berkeley/\case/{{contour_adaptive_std_noise_0.00}}}&%
\includegraphics[height=\fH]{segmentation/berkeley/\case/{{contour_adaptive_std_noise_0.40}}}&%
\includegraphics[height=\fH]{segmentation/berkeley/\case/{{contour_adaptive_std_noise_0.80}}}&%
\includegraphics[height=\fH]{segmentation/berkeley/\case/{{contour_adaptive_std_noise_1.20}}}&%
\includegraphics[height=\fH]{segmentation/berkeley/\case/{{contour_adaptive_std_noise_1.60}}}&%
\includegraphics[height=\fH]{segmentation/berkeley/\case/{{contour_adaptive_std_noise_2.00}}}\\[-\dp\strutbox] 
original & $\sigma = 0.4$ & $\sigma = 0.8$ & $\sigma = 1.2$ & $\sigma = 1.6$ & $\sigma = 2.0$
\end{tabular}
\caption{[Segmentation] Visual comparison of the segmentation results with the best F-measure. Yellow line indicates the ground truth.
(top) input images with spatially biased Gaussian noises with different standard deviations. 
(middle) optimal solutions using the constant regularity.
(bottom) optimal solutions using our adaptive regularity.}
\label{fig:segmentation_image}
\end{figure*}
%
%
\def\fH{120pt}
\def\case{353013}
\begin{figure*}[bt]
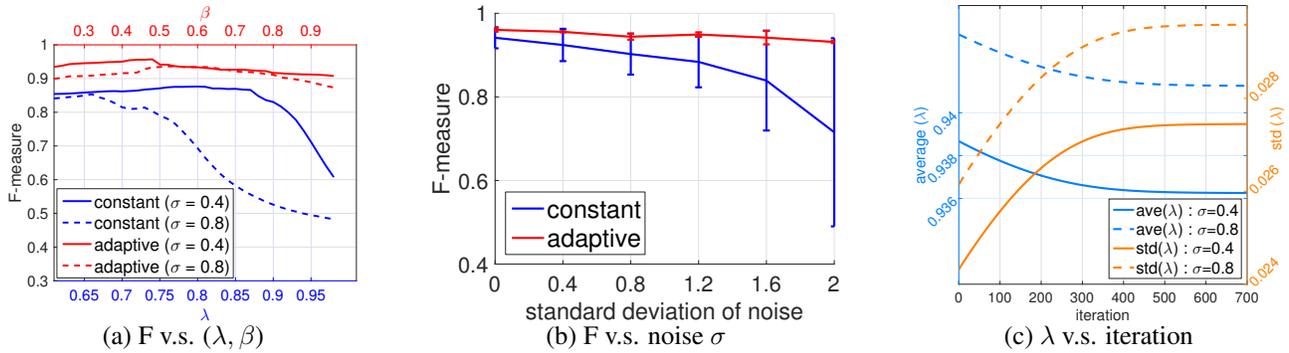

\begin{center}
\begin{tabular}{c@{ }c@{ }c}
\includegraphics[height=\fH]{segmentation/berkeley/\case/{{measure_vs_params_from_std_noise_0.40}}} \hspace{10pt} & \hspace{10pt} %
\includegraphics[height=\fH]{segmentation/berkeley/\case/{{errorbar_F_using_minmax}}} \hspace{10pt} & \hspace{10pt} %
\includegraphics[height=\fH]{segmentation/berkeley/\case/{{lambda_vs_iteration}}}\\[-\dp\strutbox]
(a) F v.s. ($\lambda, \beta$) \hspace{10pt} & \hspace{10pt} (b) F v.s. noise $\sigma$ \hspace{10pt} & \hspace{10pt} (c) $\lambda$ v.s. iteration
\end{tabular}
\end{center}
\caption{[Segmentation] (a) F-measures with varying regularization parameters, $\lambda$ (constant) and $\beta$ (adaptive). (b) F-measures with respect to varying noise standard deviations. (c) Average (left y-axis) and standard deviation (right y-axis) of our adaptive regularity $\lambda$ with respect to the optimization iteration.}
\label{fig:segmentation_error_param}
\end{figure*}
%
%
The quantitative evaluation for the denoising is also presented based on (a) PSNR and (b) SSIM in Fig.~\ref{fig:denoise_error_param} where each measure is computed with varying regularity parameters, $\lambda$ for the constant regularity (bottom x-axis) and $\beta$ for our adaptive one (top x-axis), for the example images with noise levels $\sigma = 0.2$ and $\sigma = 0.4$ in Fig.~\ref{fig:denoise_image}. It is shown that the both performance measures with the adaptive regularity parameter are consistently better over the constant one across the entire range of the parameters. The overall errors, PSNR (left y-axis) and SSIM (right y-axis), for the images used in the segmentation experiments presented in the following section are shown in (c) where the adaptive regularity parameter is superior to the constant one for all the noise levels. 

%
\subsection{Image Segmentation}

We solve the energy minimization problem in~\eqref{eq:energy_const} for the image segmentation of noisy images with spatially biased Gaussian noises of different levels using Berkeley segmentation dataset~\cite{MartinFTM01} from which images that are suited for the bi-partitioning segmentation model are only selected for the illustrative objective. We use F-measure for the quantitative evaluation of the segmentation result. In Fig.~\ref{fig:segmentation_image}, the input images with different noise standard deviations $\sigma$ are shown (top), and the segmentation results using the constant regularization parameter (middle) and our adaptive one (bottom) are shown in blue and red, respectively. The ground truth boundary\footnote{The ground truth boundary is not uniquely provided in the Berkeley dataset and we have chosen the one suited for our bi-partitioning segmentation model.} is indicated in yellow. The graphical illustration of the segmentation results demonstrate that the constant regularization parameter suffers from the spatially varying noise levels in which large regularity due to the locally present high noise levels excessively blurs the segmenting boundary at the regions with low noise levels, and in the same way small regularity due to the locally present low noise levels unnecessarily captures undesirable insignificant details.
In Fig.~\ref{fig:segmentation_error_param}, the quantitative evaluation for the segmentation is presented based on F-measure with respect to (a) the varying regularization parameters, $\lambda$ for the constant regularization (bottom x-axis) and $\beta$ for our adaptive one (top x-axis), and (b) the different noise levels $\sigma$ using the example images that are suited for the bi-partitioning model. It is shown that our adaptive regularization parameter consistently outperforms the constant one over both the entire range of parameters and all the noise levels. The temporal adjustment of our adaptive regularization to the local property of observation is demonstrated in (c) where the average and the standard deviation of the values in $\lambda$ for our adaptive parameter are plotted over the optimization iteration. It is shown that the average of adaptive $\lambda$ decreases and its standard deviation increases gradually over iterations until convergence since the initialization for the solution is made with the original input image.  
%
%
%
%
%
%
%

\subsection{Motion Estimation}
\def\gap{54pt}
\def\fH{115pt}
\def\fW{80pt}
\def\tW{33pt}
\begin{table*}[htb]
	\centering
	\scriptsize
	\begin{tabular}{|r|p{\tW}|p{\tW}|p{\tW}|p{\tW}|p{\tW}|p{\tW}|p{\tW}|p{\tW}|}
		\hline
		& \multicolumn{2}{c|}{Static $\lambda=0.01$} & \multicolumn{2}{c|}{Static $\lambda=0.2$} & \multicolumn{2}{c|}{Best static} & \multicolumn{2}{c|}{Adaptive $\beta$=1}\\
		\hline
		\textbf{Dataset:} & \textbf{AEE:} & \textbf{AE:} & \textbf{AEE:} & \textbf{AE:} & \textbf{AEE:} & \textbf{AE:} & \textbf{AEE:} & \textbf{AE:} \\
		\hline
		\begin{filecontents*}{tableResults.csv}
Var1,Var2,Var3,Var4,Var5,Var6,Var7,Var8,Var9
Dimetrodon,0.09,0.019,0.15,0.032,0.082,0.018,0.082,0.017
Grove2,0.185,0.028,0.108,0.019,0.082,0.014,0.089,0.016
Hydrangea,0.185,0.023,0.122,0.014,0.088,0.012,0.096,0.013
Urban2,0.445,0.027,0.509,0.062,0.21,0.018,0.214,0.018
Venus,0.64,0.051,0.291,0.059,0.214,0.042,0.206,0.044
\end{filecontents*}
\csvreader[late after line=\\\hline]{tableResults.csv}{}{\csvcoli &\csvcolii & \csvcoliii & \csvcoliv & \csvcolv & \csvcolvi & \csvcolvii & \csvcolviii & \csvcolix}%
	\end{tabular}
	\caption{Endpoint error and angular error for static and adaptive regularization parameters. The adaptive case does not necessarily create the best result (compared to static choices), but yields encouraging results along all datasets.}
	\label{parametersAndRanks}
\end{table*}
%
%
The quantitative evaluation of the algorithm is performed using the angular error (AE)~\cite{barron1994performance} and absolute endpoint error (EE)~\cite{otte1994optical}. The absolute endpoint error is computed by the Euclidean distance between the calculated motion field and a given ground truth vector field. For the angular error, first ground-truth and calculated vector field are projected into the three-dimensional space. Afterwards, the average angle between both fields is calculated.
A set of gray-valued images with given ground truth flow from the Middlebury optical flow database~\cite{baker2011database} has been used to quantify the proposed adaptive regularization strategy against the static one.
\begin{figure}
	\centering
	\includegraphics[width=0.23\textwidth]{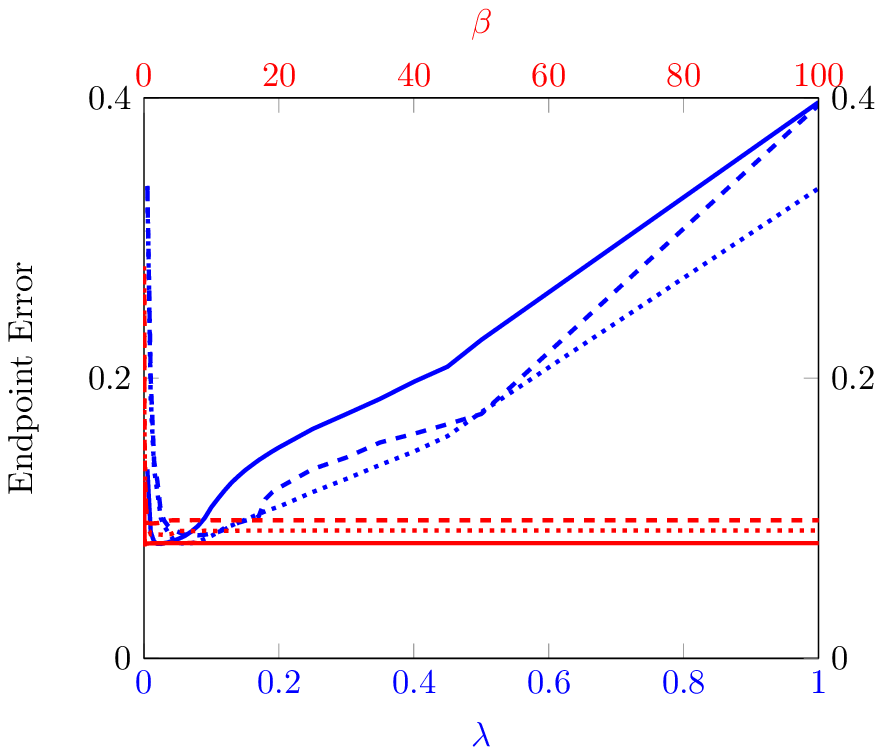}
	\includegraphics[width=0.233\textwidth]{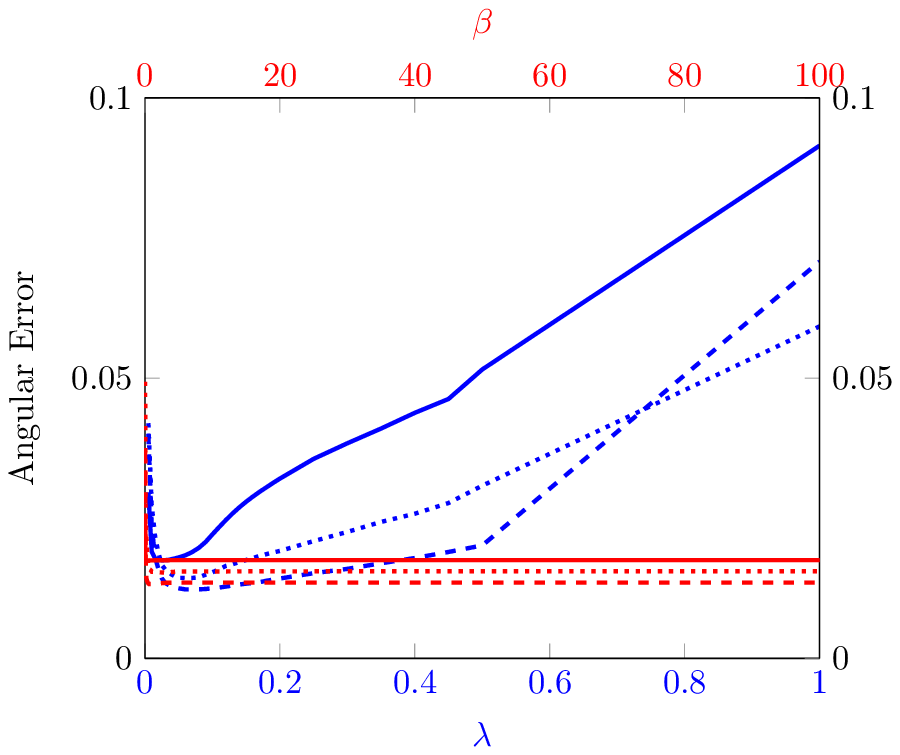}
	\caption{Left: Endpoint error (left) and angular error (right) with respect to $\lambda$ (blue) and $\beta$ (red). The Dimetrodon (solid), Grove2 (dotted) and Hydrangea (dashed) datasets from \cite{baker2011database} were used to generate these graphs.}
	\label{figureMotionRobust}
\end{figure}
First of all, the robustness of our method is shown in Fig.~\ref{figureMotionRobust}. The angular and absolute endpoint errors for a wide range of static regularization parameters $\lambda$ and weights $\beta$ for the  adaptive strategy is plotted. The graph indicates the robustness of the adaptive strategy, except for very small values of $\beta$. In general, there exist static parameters $\lambda$ that creates equally good results, but those are unstable with respect to the chosen dataset.
Table~\ref{parametersAndRanks} includes a quantitative performance overview of our method. Angular and endpoint error are listed for static regularization parameters $\lambda=0.01$ and $\lambda=0.2$, and for the best static parameter $\lambda$ that could be found in $[0,1]$. In addition, we list the errors for the simple choice $\beta=1$. The results demonstrate that the adaptive strategy may not necessarily generate the overall best result compared to static parameters, but yields nearly equally good results without any tuning of parameters.
Finally, a visual comparison of the generated velocity fields is provided in Fig.~\ref{figureMotionResults} where the estimated flows obtained with best parameters and their ground truth are presented.
\def\fW{50pt}
\begin{figure}[htb]
	\centering
	\begin{tabular}{c@{ }c@{ }c@{ }c@{ }c}
		\includegraphics[width=\fW]{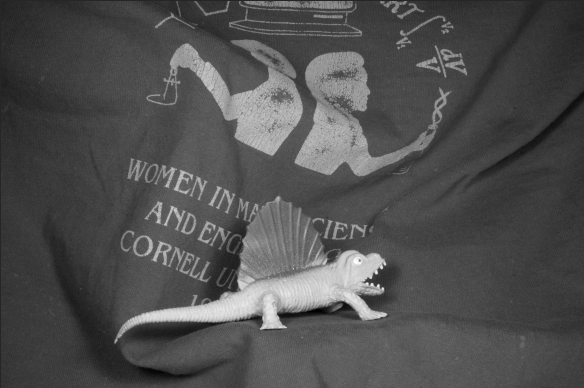}&
		\includegraphics[width=\fW]{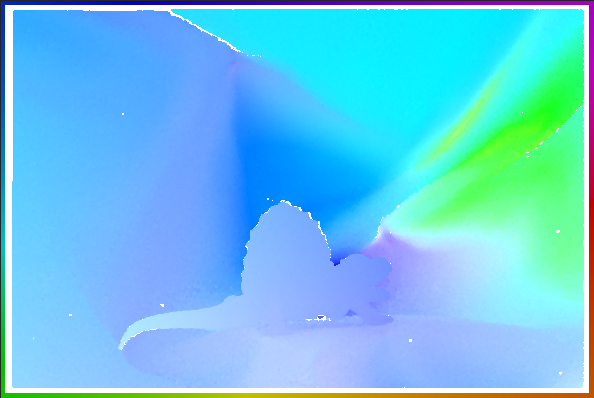}&
		\includegraphics[width=\fW]{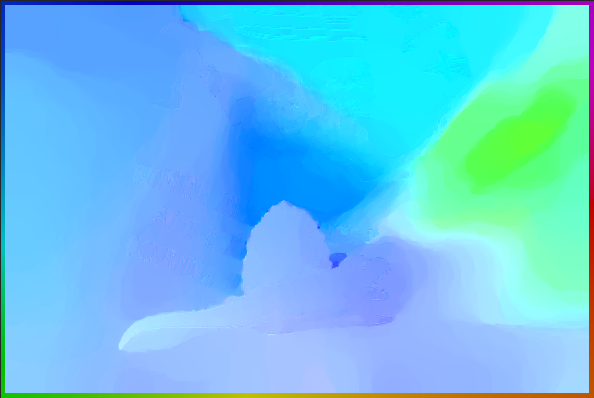}&
		\includegraphics[width=\fW]{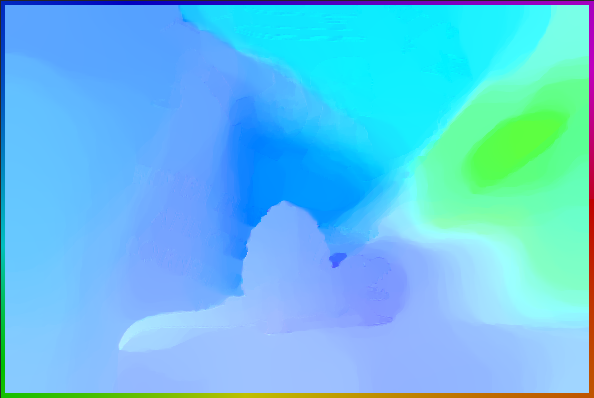}\\[-\dp\strutbox] 
		\includegraphics[width=\fW]{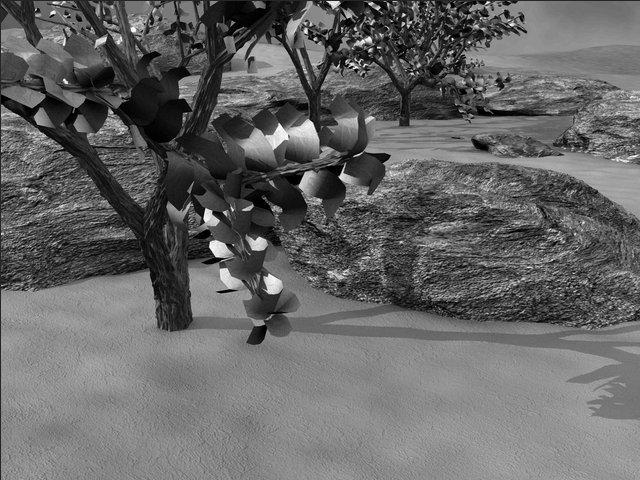}&
		\includegraphics[width=\fW]{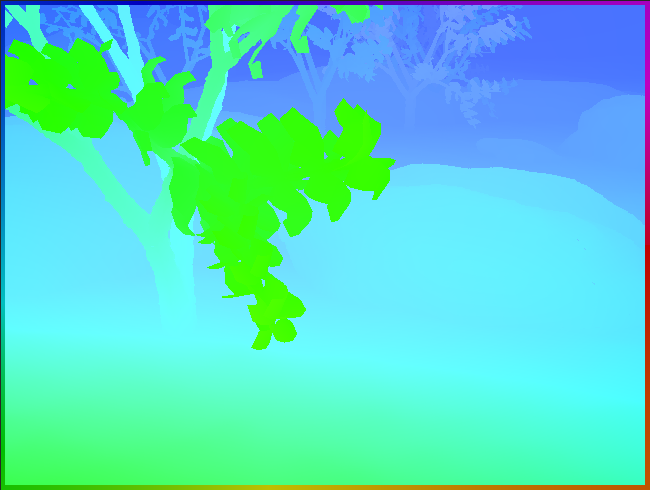}&
		\includegraphics[width=\fW]{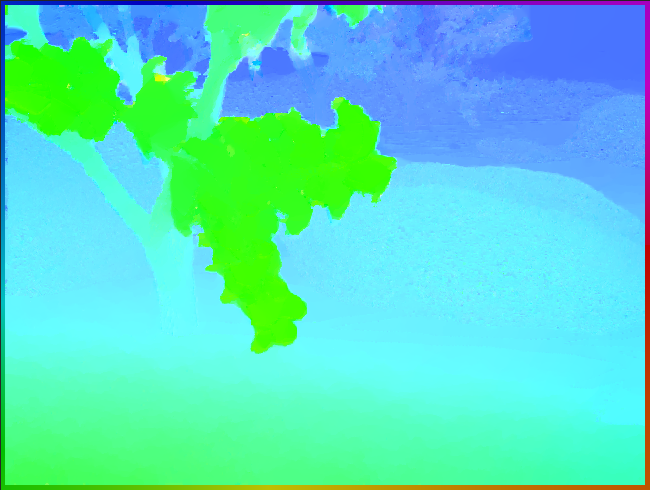}&
		\includegraphics[width=\fW]{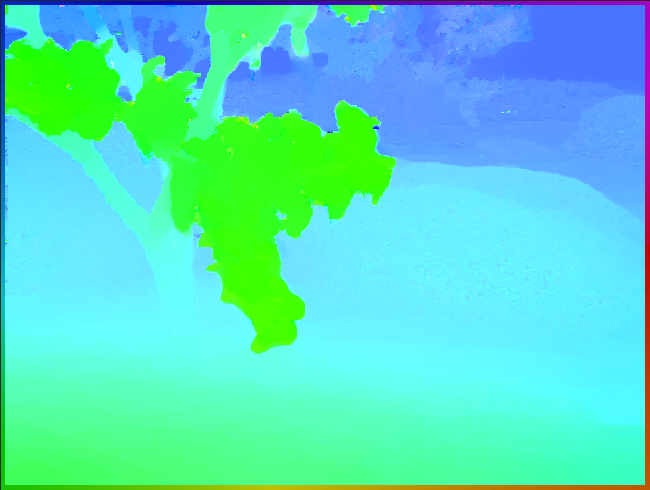}\\[-\dp\strutbox]
		\includegraphics[width=\fW]{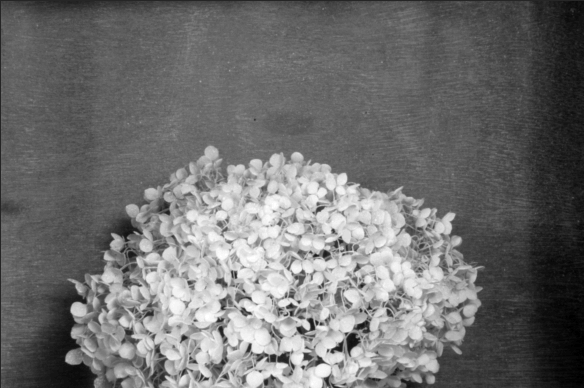}&
		\includegraphics[width=\fW]{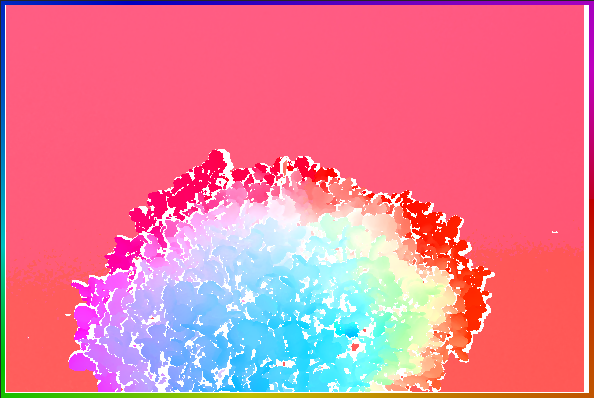}&
		\includegraphics[width=\fW]{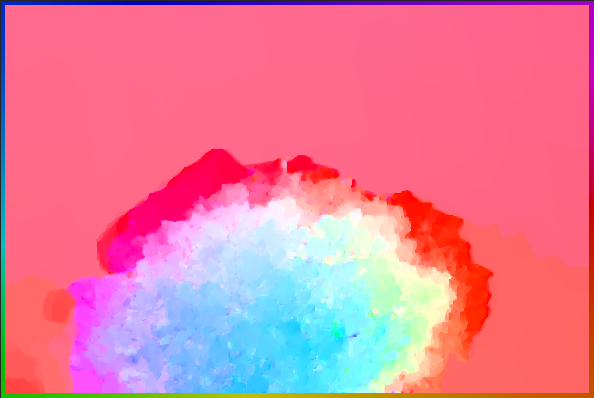}&
		\includegraphics[width=\fW]{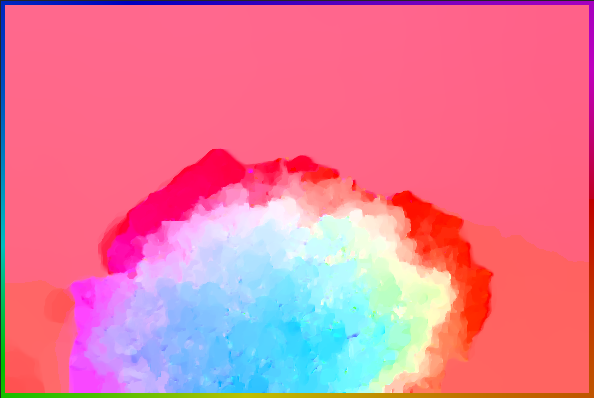}\\[-\dp\strutbox]
		\includegraphics[width=\fW]{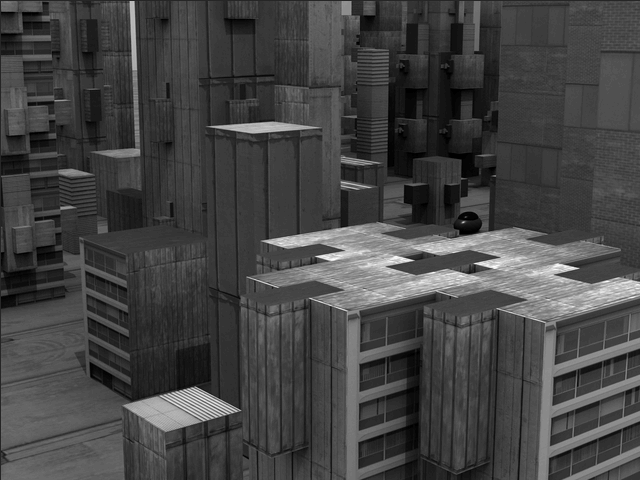}&
		\includegraphics[width=\fW]{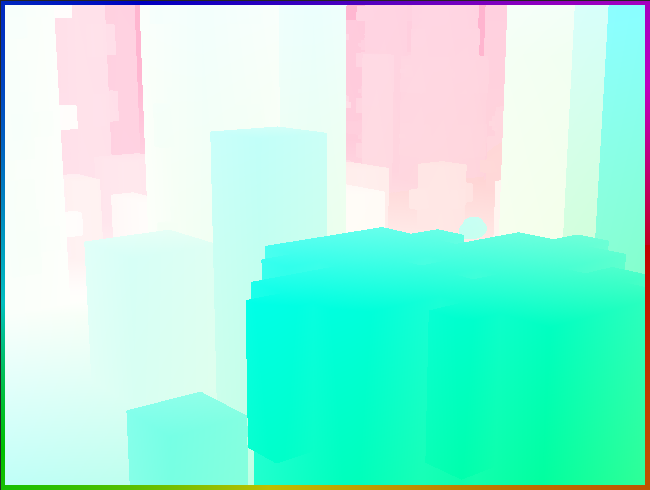}&
		\includegraphics[width=\fW]{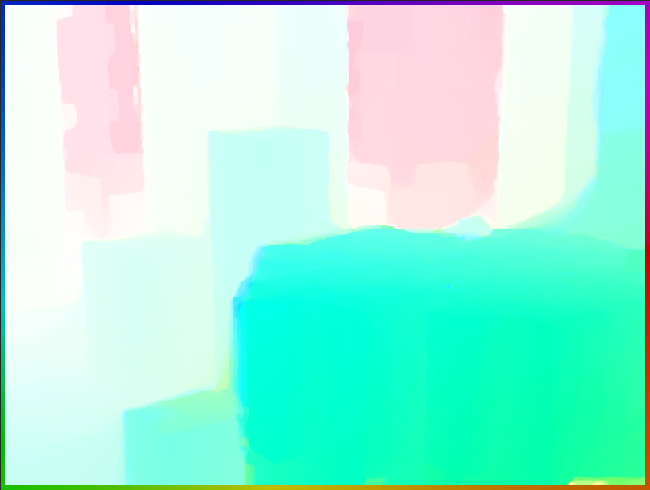}&
		\includegraphics[width=\fW]{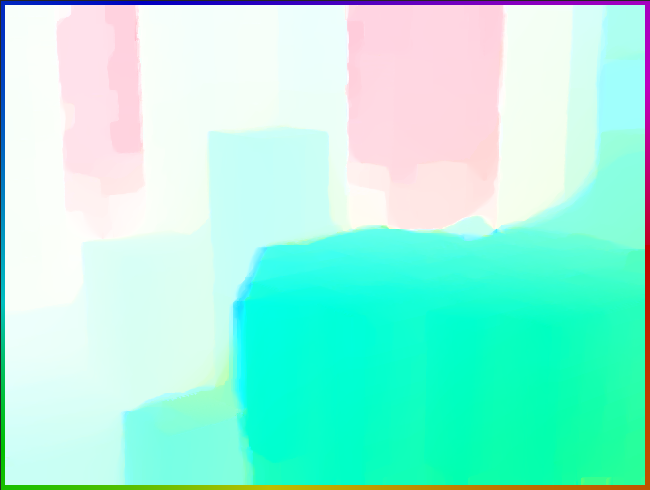}\\[-\dp\strutbox]
		\includegraphics[width=\fW]{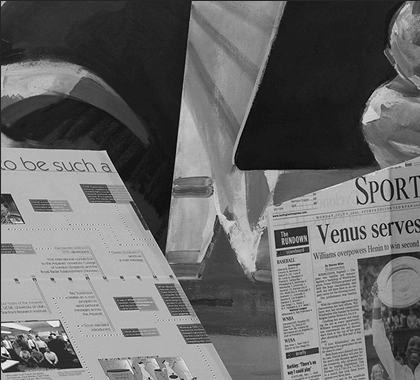}&
		\includegraphics[width=\fW]{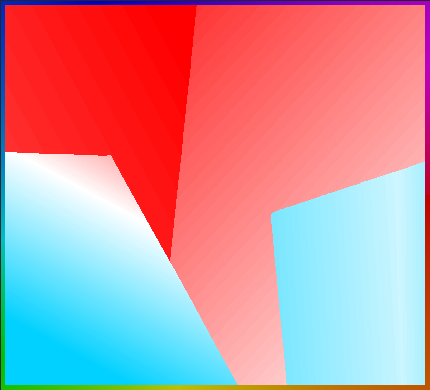}&
		\includegraphics[width=\fW]{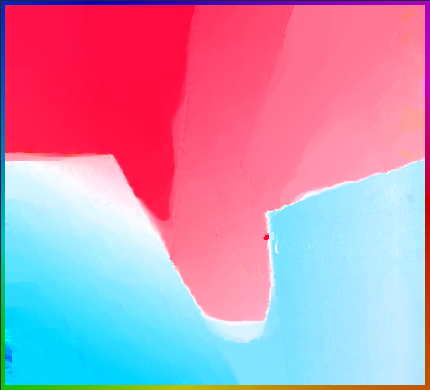}&
		\includegraphics[width=\fW]{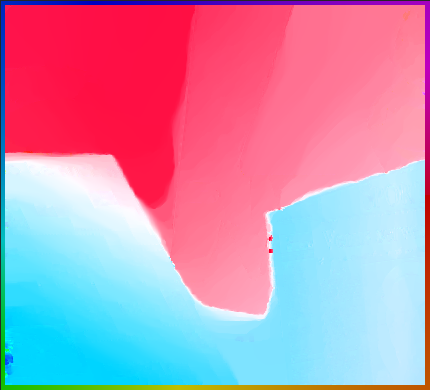}
	\end{tabular}
	\caption{Visual illustration of the optical flow. Left to right: Image, ground truth flow field, result with adaptive parameter choice, best result with static parameter. The velocity is represented by the conventional color coding scheme.}
	\label{figureMotionResults}
\end{figure}
%
%
\section{Conclusion and Discussion}

We have presented a novel regularization scheme in a variational framework where the energy functional mainly consists of data fidelity, regularization and a control parameter for their trade-off.
In the energy optimization procedure, the relative weight between data fidelity and regularization is determined based on the residual that measures how well the observed data fits to the model under consideration.
We have applied our proposed algorithm to the classical imaging problems including denoising, segmentation and motion estimation, and we also have presented their optimization algorithms based on the ADMM method.
The adaptive regularization scheme has been shown to be more effective in particular when the distribution of degrading factors such as noise is spatially biased. The parameter involved in our adaptive regularization scheme has been shown to be robust in contrast to the constant regularization parameter that is sensitive and often critical to the quality of solution. 
Experimental results for each imaging task have been obtained based on benchmark dataset and it has been shown that higher accuracy is achieved for each imaging task based on its classical energy functional merely by replacing the constant regularization with our adaptive one. 

\newpage

\begin{appendix}
	\section{Appendix}
		\subsection{Proofs}
		
		\subsubsection*{Proof of Lemma 1}
		
		Let $u^*$ satisfy $\rho(u^*)$, hence $\lambda \equiv 1$. Then 
		$$E_{\lambda(u^*)}(u) = \int_\Omega \rho(\textup{?}) \ud x, $$
		which is obviously minimized by $u^*$. 
		
		\subsubsection*{Proof of Theorem 1}
		
		We provide a sketch of the fixed point argument in the following.  The topology we use is strong convergence of $(u,\rho(u))$ in $L^1(\Omega)\times L^1(\Omega)$, and we construct a self-map on this space. Then trivially the map to $ \rho(u) \in L^1(\Omega)$ is continuous.
		With the properties of the convolution kernel $G$ we immediately see that $\rho \in L^1(\Omega) \mapsto G*\rho \in C^1(\Omega)$ is continuous and compact. Moreover, the map $ G*\rho \in C^1(\Omega) \mapsto \lambda \in C^1(\Omega)$ is continuous. Finally we see from a standard continuous dependence argument on the variational problem that $\lambda \in C^1(\Omega) \mapsto (u,\rho(u)) \in
		BV_0(\Omega) \times L^1(\Omega)$ is continuous, and the 
		 continuous embedding of $BV_0(\Omega)$ into $L^1(\Omega)$ finally implies the continuity and compactness and fixed point operator on these spaces. In order to apply a Schauder's fixed-point theorem and conclude the existence of a fixed point, it suffices to show that some bounded set is mapped into itself. For this sake let $C_0 = \int_\Omega \rho(0) \ud x$, and choose $c$ such that
		$$ (1-\epsilon) \exp\left(-\frac{\Vert G\Vert_\infty C_0}{\beta c}\right) \geq c. $$
		The existence of such a $c$ is guaranteed for $\beta$ sufficiently large. Now let 
		$$ \int \rho(u) \ud x \leq \frac{C_0}c, $$ then we obtain with a standard estimate of the convolution and monotonicity of the exponential function that
		$$ 1 \geq \lambda = (1-\epsilon) \exp\left(-\frac{G*\rho}\beta\right) \geq c. $$
		Moreover, there exists a constant $\tilde c$ such that $1 \geq 1-\lambda \geq \tilde c$. Hence, a minimizer $u$ of $E_\lambda$ satisfies
		\begin{align*} c \int_\Omega & \rho(u) \ud x + \tilde c \int_\Omega |\nabla u| \ud x \\
		& \leq E_\lambda(u) \leq E_\lambda(0) \leq \int_\Omega \rho(0) \ud x = C_0.
		\end{align*}
		Hence using the closed set of $u,\rho$ such that 
		$$ \Vert \rho \Vert_{L^1} \leq \frac{C_0}{c}, \qquad \vert u \vert_{BV} \leq \frac{C_0}{\tilde c}, $$
		we obtain a self-mapping by our fixed-point operator.
		
		\subsection{Optimization}
		
		\subsubsection{Optimization for Image Denoising}
		
		%
		The associated augmented Lagrangian with augmentation parameter $\mu > 0$ for the splitting of the variables in the scaled form reads:
		\begin{align*} 
		\mathcal{L}_{\mu}(u, z, y) = \langle \lambda, & \frac{1}{2} (u - f)^2 \rangle + \langle 1 - \lambda, \| z \|_1 \rangle \\
		&+ \frac{\mu}{2} \| \nabla u - z + y \|_2^2,
		\end{align*}
		where $y$ is a Lagrangian multiplier associated with $u$ and $z = \nabla u$. 
		%
		%
		The optimality condition of the update for the primal variable $u^{k+1}$ with the linearisation at around $u^k$ yields:
		%
		\begin{align*}
		u^{k+1} =& \argmin_u \langle \lambda^k, \frac{1}{2} (u - f)^2 \rangle\\
		&+ \mu \nabla^T ( \nabla u^k - z^k + y^k ) u + \frac{\tau}{2} \| u - u^k \|_2^2 
		\end{align*}
		where $- \nabla^T$ represents the discrete divergence operator. 
		Then, the solution for updating $u$ is obtained by: 
		\begin{align*}
		u^{k+1} &=\frac{1}{\lambda + \tau} \left( \tau u^k + \lambda f - \mu \nabla^T \left( \nabla u^k - z^k + y^k \right) \right). \label{solution:u_denoise}
		\end{align*}
		The optimality condition of the update for the variable $z^{k+1}$ yields:
		\begin{align}
		0 &\in (1 - \lambda^k) \partial \| z \|_1 + \mu (\nabla u^{k+1} - z + y^k).
		\end{align}
		%
		Then, the solution for updating $z$ is obtained by:
		\begin{align*}
		z^{k+1} 
		&=\prox \left( \nabla u^{k+1} + y^k \left| \frac{1 - \lambda^k}{\mu} \gamma \right. \right)\\
		&= S\left( \nabla u^{k+1} + y^k \left| \frac{1 - \lambda^k}{\mu} \right. \right). 
		\end{align*}
		

		\subsubsection{Optimization for Image Segmentation}
		
		An alternative minimisation is performed with respect to $u$ given fixed $c_1$ and $c_2$, and then with respect to $c_1$ and $c_2$ given a fixed $u$. For a fixed $u$, the optimal $c_1$ and $c_2$ can be obtained by:
		\begin{align*}
		c_1 = \frac{\int_{\Omega} f u \ud x}{\int_{\Omega} u \ud x}, \quad  
		c_2 = \frac{\int_{\Omega} f (1 - u) \ud x}{\int_{\Omega} (1 - u) \ud x}.
		\end{align*}
		%
		The associated augmented Lagrangian with augmentation parameter $\mu > 0$ for the splitting of the variables in the scaled form reads:
		\small
		\begin{align*} 
		\mathcal{L}_{\mu}(u, z, y) = &\langle \lambda, \rho(u) \rangle + \langle 1 - \lambda, \gamma(z) \rangle \\
		&+ \frac{\mu}{2} \| \nabla u - z + y \|_2^2 + \delta_{\mathcal{C}} (u),
		\end{align*}
		where $y$ is a Lagrangian multiplier associated with $u$ and $z = \nabla u$. 
		%
		The optimality condition of the update for the primal variable $u^{k+1}$ with the linearisation at around $u^k$ yields:
		{\footnotesize
		\begin{align*}
		&u^{k+1}\\
		&=\argmin_u \langle \lambda^k, \rho(u) \rangle + \mu \nabla^T ( \nabla u^k - z^k + y^k ) u + \frac{\tau}{2} \| u - u^k \|_2^2 + \delta_{\mathcal{C}} (u)\\
		&=\argmin_u \frac{\tau}{2} \left\| u - \left( u^k - \frac{\lambda^k}{\tau} q -\frac{\mu}{\tau} \nabla^T \left( \nabla u^k - z^k + y^k \right) \right) \right\|_2^2 + \delta_{\mathcal{C}} (u) 
		\end{align*}
		}%
		where $q = (f - c_1)^2 - (f - c_2)^2$.
		Then, the solution for updating $u$ is obtained by: 
		{\footnotesize
		\begin{align*}
		&u^{k+1}\\
		&=\Pi_{\mathcal{C}} \left( u^k - \frac{\lambda^k}{\tau} \left( (f - c_1)^2 - (f - c_2)^2 \right) -\frac{\mu}{\tau} \nabla^T \left( \nabla u^k - z^k + y^k \right) \right) 
		\end{align*}
		}
		%
		where the projection operator $\Pi_{\mathcal{C}}(u) = \argmin_{v \in \mathcal{C}} \| u - v \|_2$.
		The optimality condition of the update for the variable $z^{k+1}$ yields:
		\begin{align}
		0 &\in (1 - \lambda^k) \partial \gamma(z) + \mu (\nabla u^{k+1} - z + y^k).
		\end{align}
		%
		Then, the solution for updating $z$ is obtained by:
		\begin{align}
		z^{k+1} &=\prox \left( \nabla u^{k+1} + y^k \left| \frac{1 - \lambda^k}{\mu} \gamma \right. \right) \\
		&= S\left( \nabla u^{k+1} + y^k \left| \frac{1 - \lambda^k}{\mu} \right. \right). \label{eq:shrink_z}
		\end{align}

		%
		\subsubsection{Optimization for Motion Estimation}
		
		The associated augmented Lagrangian with augmentation parameter $\mu > 0$ for the splitting of the variables in the scaled form reads:
		\begin{align} \label{eq:lagrange_motion}
		\mathcal{L}_{\mu}&(u, v, y, z, s, p, q, r) = \langle \lambda, \gamma(s) \rangle + \langle 1 - \lambda, \gamma(y) \rangle \\
		&+ \langle 1 - \lambda, \gamma(z) \rangle + \frac{\mu}{2} \| \nabla u - y + p \|_2^2 \nonumber \\
		&+ \frac{\mu}{2} \| \nabla v - z + q \|_2^2 + \frac{\mu}{2} \| I_t + \nabla I \cdot w - s + r \|_2^2, 
		\end{align}
		where $p$ is a Lagrangian multiplier associated with $u$ and $y = \nabla u$, $q$ is with $v$ and $z = \nabla v$, and $r$ is with $w$ and $s = I_t + \nabla I \cdot w$.
		The optimality condition of the update for the primal variables $u^{k+1}$ and $v^{k+1}$ with the linearisation at around $u^k$ and $v^k$ yields:
		\begin{align*}
		u^{k+1} =\argmin_u &\frac{\mu}{2} \| I_t + \nabla I \cdot w - s + r \|_2^2 \\
		& + \mu \nabla^T ( \nabla u^k - y^k + p^k ) u + \frac{\tau}{2} \| u - u^k \|_2^2,
		\end{align*}
		\begin{align*}
		v^{k+1} =\argmin_v &\frac{\mu}{2} \| I_t + \nabla I \cdot w - s + r \|_2^2 \\
		&+ \mu \nabla^T ( \nabla v^k - z^k + q^k ) u + \frac{\tau}{2} \| v - v^k \|_2^2.
		\end{align*}
		%
		The optimality condition of the update and the solution for the variables $y^{k+1}, z^{k+1}$ and $s^{k+1}$ yield:
		\begin{align*}
		0 &\in (1 - \lambda^k) \partial \gamma(y) - \mu (\nabla u^{k+1} - y + p^k), \\
		y^{k+1} & =\prox \left( \nabla u^{k+1} + p^k \left| \frac{1 - \lambda^k}{\mu} \gamma \right. \right),\\
		0 &\in (1 - \lambda^k) \partial \gamma(z) - \mu (\nabla v^{k+1} - z + q^k), \\
		 z^{k+1} & =\prox \left( \nabla v^{k+1} + q^k \left| \frac{1 - \lambda^k}{\mu} \gamma \right. \right),\\
		0 &\in \lambda^k \partial \gamma(s) - \mu (I_t + \nabla I \cdot w - s + r), \\
		s^{k+1} & =\prox \left( I_t + \nabla I \cdot w + r \left| \frac{\lambda^k}{\mu} \gamma \right. \right).
		\end{align*}
		%
		%
		
		
\end{appendix}

\newpage
{\small
\bibliographystyle{ieee}
\bibliography{convex}
}

\end{document}